\documentclass[letterpaper, 10 pt, conference]{ieeeconf}
\IEEEoverridecommandlockouts 
\usepackage[backend=bibtex,
            hyperref=true,
            url=false,
            isbn=false,
            doi=false,
            backref=false,
            style=ieee,
            natbib=true,
            mincitenames=1,
            maxcitenames=1,
            citestyle=numeric-comp,
            sorting=nyt,
            block=none]{biblatex}
        
\usepackage{url}

\usepackage{hyperref}

\usepackage{amsmath,amssymb}
\usepackage{bm}

\numberwithin{equation}{section} 

\usepackage{graphicx}
\graphicspath{ {figures/} }
\usepackage{multicol,multirow}

\usepackage{tabu}
\usepackage{booktabs}
\usepackage{caption}

\usepackage[dvipsnames]{xcolor}
\usepackage{autolabtools}

\title{\LARGE \bf
X-Ray: Mechanical Search for an Occluded Object \\
by Minimizing Support of Learned Occupancy Distributions
}

\author{Michael Danielczuk$^{1}$, Anelia Angelova$^{2}$, Vincent Vanhoucke$^{2}$, Ken Goldberg$^{1}$%
\thanks{$^{1}$The Autolab at University of California, Berkeley. $^{2}$Robotics at Google. mdanielczuk@berkeley.edu, anelia@google.com, vanhoucke@google.com, goldberg@berkeley.edu}
}

\begin{document}

\maketitle


\begin{abstract}

For applications in e-commerce, warehouses, healthcare, and home service, robots are often required to search through heaps of objects to grasp a specific target object. For mechanical search, we introduce X-Ray, an algorithm based on learned occupancy distributions. We train a neural network using a synthetic dataset of RGBD heap images labeled for a set of standard bounding box targets with varying aspect ratios. X-Ray minimizes support of the learned distribution as part of a mechanical search policy in both simulated and real environments. We benchmark these policies against two baseline policies on 1,000 heaps of 15 objects in simulation where the target object is partially or fully occluded. Results suggest that X-Ray is significantly more efficient, as it succeeds in extracting the target object 82\% of the time, 15\% more often than the best-performing baseline. Experiments on an ABB YuMi robot with 20 heaps of 25 household objects suggest that the learned policy transfers easily to a physical system, where it outperforms baseline policies by 15\% in success rate with 17\% fewer actions. Datasets, videos, and experiments are available at \url{https://sites.google.com/berkeley.edu/x-ray}.

\end{abstract}
\section{Introduction} \label{sec:introduction}
Mechanical search -- extracting a desired object from a heap of objects -- is a fundamental task for robots in unstructured e-commerce warehouse environments or for robots in home settings. It remains challenging due to uncertainty in perception and actuation as well as lack of models for occluded objects in the heap. 



Data-driven methods are promising for grasping unknown objects in clutter and bin picking~\cite{mahler2019learning,pinto2016supersizing,kalashnikov2018qt,morrison2018closing,gualtieri2016high}, and can reliably plan grasps on the most accessible object without semantic knowledge of the target object. Some reinforcement learning~\cite{yang2019deep,jang2017end} or hierachical~\cite{danielczuk2019mechanical} mechanical search policies use semantics, but have so far been limited to specific objects or heuristic policies. 

\begin{figure}[t!]
\centering
\includegraphics[width=\linewidth]{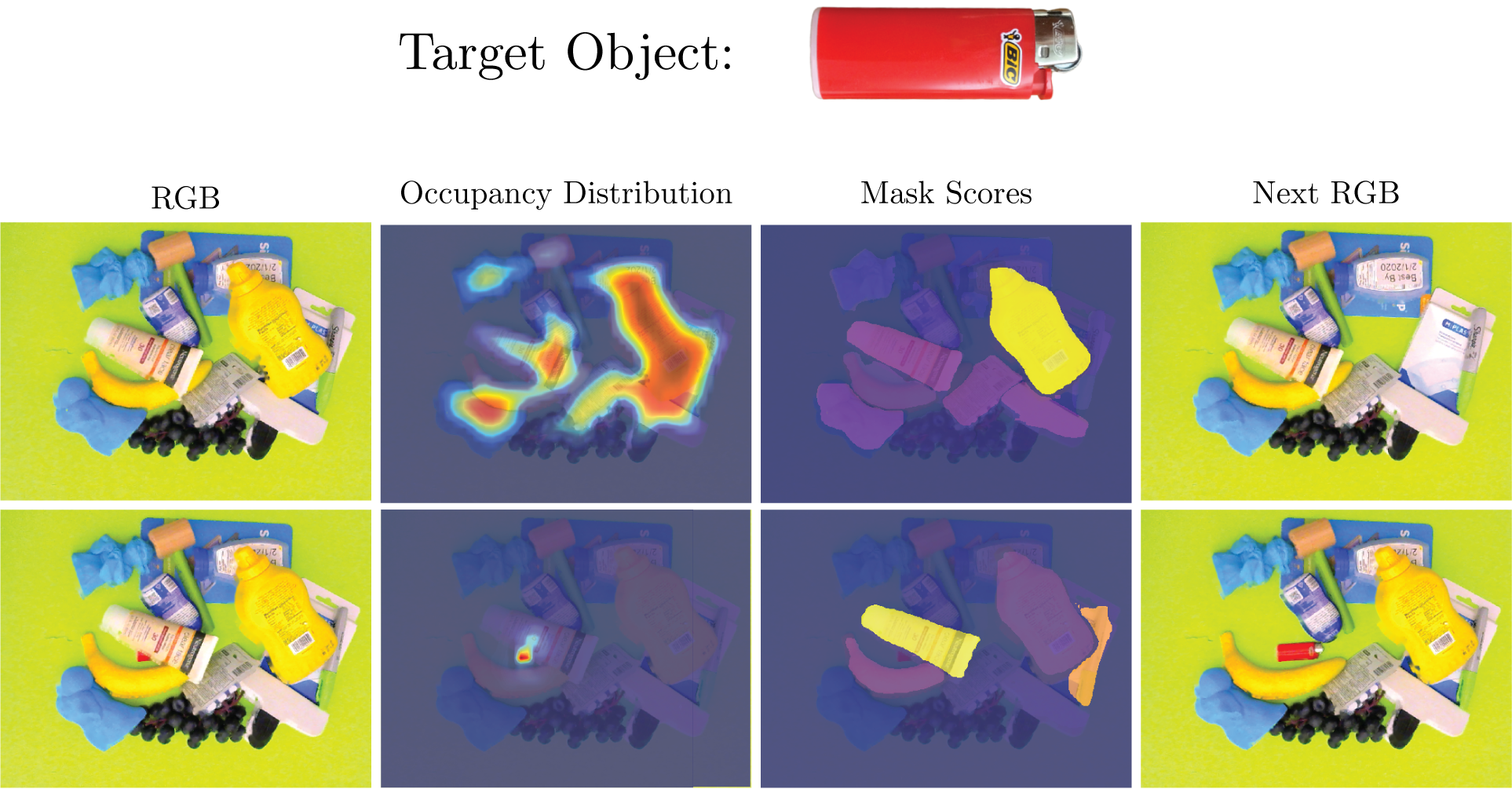}
\caption{Mechanical search with a fully occluded target object (top row) and a partially occluded target object (bottom row). We predict the target object occupancy distribution, which depends on the target object's visibility and the heap (second column). Each pixel value in the distribution image corresponds to the likelihood of that pixel containing part of the target object. X-Ray plans a grasp on the object that minimizes the estimated support of the resulting occupancy distribution to minimize the number of actions to extract the target object. We show two nearly-identical heaps; in the fully occluded case, X-Ray grasps the mustard bottle whereas in the partially occluded case, the policy grasps the face lotion (third column), resulting in the respective next states (fourth column).}
\label{fig:splash}
\vspace{-8pt}
\end{figure}

In this paper, we draw on recent work on shape completion to reason about occluded objects~\cite{varley2017shape,price2019inferring} and work on predicting multiple pose hypotheses~\cite{manhardt2018explaining,rupprecht2017learning}. X-Ray combines occlusion inference and hypothesis predictions to estimate an occupancy distribution for the bounding box most similar to the target object to estimate likely poses -- translations and rotations in the image plane. X-Ray can efficiently extract the target object from a heap where it is fully occluded or partially occluded (Figure~\ref{fig:splash}).

This paper provides four contributions:
\begin{enumerate}
\item X-Ray (maXimize Reduction in support Area of occupancY distribution): a mechanical search policy that minimizes support of learned occupancy distributions.
\item An algorithm for estimating target object occupancy distributions using a set of neural networks trained on a dataset of synthetic images that transfers seamlessly to real images.
\item A synthetic dataset generation method and 100,000 RGBD images of heaps labeled with occupancy distributions for a single partially or fully occluded target object, constructed for transfer to real images.
\item Experiments comparing the mechanical search policy against two baselines in 1,000 simulated and 20 physical heaps that suggest the policy can reduce the median number of actions needed to extract the target object by 20\% with a simulated success rate of 87\% and physical success rate of 100\%.
\end{enumerate}

\section{Related Work} \label{sec:relwork}

\subsection{Pose Hypothesis Prediction}
There is a substantial amount of related work in computer vision on 3D and 6D pose prediction of both known and unknown objects in RGB, depth, and RGBD images~\cite{kehl2017ssd,xiang2017posecnn,li2018deepim,hinterstoisser2012model}. Many of these papers assume that the target objects are either fully visible or have minor occlusions.  In addition, many assume that there is no ambiguity in object pose due to self-occlusion or rotational symmetry of the object, as these factors can significantly decrease performance for neural network-based approaches~\cite{corona2018pose}. Recent work has attempted to address the pose ambiguity that results from object geometry or occlusions by restricting the range of rotations~\cite{rad2017bb8} predicting multiple hypotheses for each detected object~\cite{rupprecht2017learning, manhardt2018explaining}. \citet{rupprecht2017learning} find that refining multiple pose hypotheses to a 6D prediction outperforms single hypothesis predictions on a variety of vision tasks, such as human pose estimation, object classification, and frame prediction. \citet{manhardt2018explaining} note that directly regressing to a rotation for objects with rotational symmetries can result in an averaging effect where the predicted pose does not match any of the possible poses; thus, they predict multiple pose hypotheses for objects with pose ambiguities to better predict the underlying pose and show Bingham distributions of the predicted hypotheses. However, only minor occlusions are considered and since ground truth pose distributions are not available for these images and objects, comparisons for continuous distributions can only be made qualitatively. Predicting multiple hypotheses or a distribution to model ambiguity has also been applied to gaze prediction from facial images~\cite{prokudin2018deep}, segmentation~\cite{kohl2018probabilistic}, and monocular depth prediction~\cite{yang2019inferring}. In contrast to these works, we learn occupancy distributions in a supervised manner.

\subsection{Object Search}
There has been a diverse set of approaches to grasping in cluttered environments, including methods that use geometric knowledge of the objects in the environment to perform wrench-based grasp metric calculations, nearest-neighbor lookup in a precomputed database, or template matching~\cite{berenson2008grasp,moll2017randomized,mahler2016dex}, as well as methods using only raw sensor data~\cite{katz2014perceiving,saxena2008learning}, commonly leveraging convolutional neural networks~\cite{kalashnikov2018qt,jang2017end,lenz2015deep}. While multi-step bin-picking techniques have been studied, they do not take a specific target object into account~\cite{mahler2017learning}.

\citet{kostrikov2016end} learn a critic-only reinforcement learning policy to push blocks in a simulated environment to uncover an occluded MNIST block. \citet{zeng2018learning} train joint deep fully-convolutional neural networks to predict both pushing and grasping affordances from heightmaps of a scene containing multicolored blocks, then show that the resulting policy (VPG) can separate and grasp novel objects in cluttered heaps. The policy can be efficiently trained on both simulated and physical systems, and can quickly learn elegant pushes to expand the set of available grasps in the scene. \citet{yang2019deep} train similar grasping and pushing networks as well as separate explorer and coordinator networks to address the exploration/exploitation tradeoff for uncovering a target object. Their policy learns to push through heaps of objects to find the target and then coordinate grasping and pushing actions to extract it, outperforming a target-centered VPG baseline in success rate and number of actions. Both approaches can generalize to objects outside the training distribution, although they are evaluated on a limited set of novel objects, and Yang \textit{et al.} separate the cases where the target object is partially occluded and fully occluded. Additionally, we focus only on grasping actions, as some mechanical search environments may be constrained or objects may be fragile.

Recently, several approaches to the mechanical search problem have been proposed, both in tabletop and bin picking environments. \citet{price2019inferring} propose a shape completion approach that predicts occlusion regions for objects to guide exploration in a tabletop scene, while \citet{xiao2019online} implement a particle filter approach and POMDP solver to attempt to track all visible and occluded objects in the scene. However, 75\% of the objects in Price \textit{et al.}'s evaluation scenes are seen in training and Xiao \textit{et al.}'s method requires models of each of the objects in the scene. We benchmark our policy on a variety of non-rigid, non-convex household objects not seen in training and require no object models. In previous work, \citet{danielczuk2019mechanical} proposed a general mechanical search problem formulation and introduced a two-stage perception and search policy pipeline. In contrast, we introduce a novel perception network and policy based on minimizing support of occupancy distributions that outperforms the methods introduced in~\cite{danielczuk2019mechanical}.
\section{Problem Statement} \label{sec:problem}
We consider an instance of the mechanical search problem where a robot must extract a known target object from a heap of unknown objects by iteratively grasping to remove non-target objects. The objective is to extract the target object using the fewest number of grasps.

\subsection{Assumptions}
\begin{itemize}
    \item One known target object, fully or partially occluded by unknown objects in a heap on a planar workspace.
    \item A robot with a gripper, an overhead RGBD sensor with known camera intrinsics and pose relative to the robot.
    \item A maximum of one object is grasped per timestep.
    \item A target object detector that can return a binary mask of visible target object pixels when queried.
\end{itemize}

\begin{figure*}[th!]
\vspace{1.5mm}
\includegraphics[width=\textwidth]{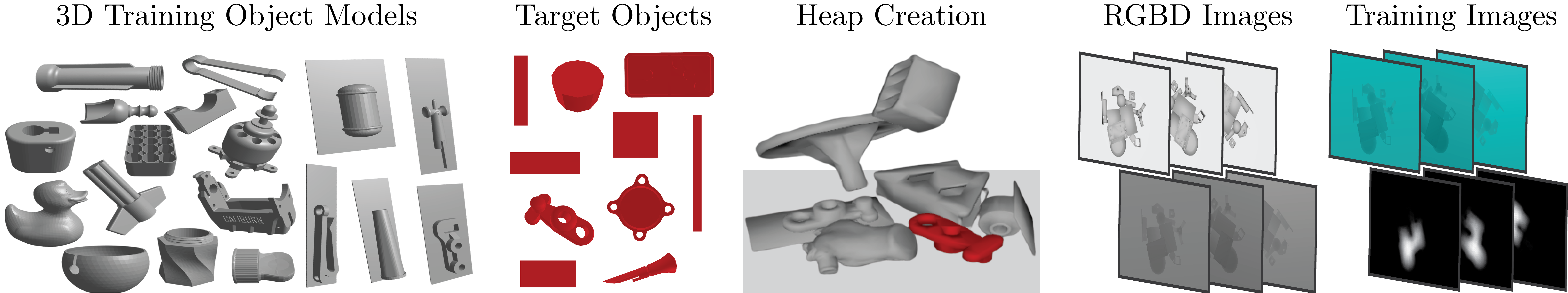}
\caption{Training dataset generation for learning the occupancy distribution function. Each dataset image is generated by sampling $N = 14$ object models from a dataset of 1296 CAD models. The target object (colored red) is dropped, followed by the $N$ other objects (colored gray), into a planar workspace using dynamic simulation. Camera intrinsics and pose are sampled from uniform distributions centered around their nominal values and an RGBD image is rendered of the scene. The augmented depth image (top right), consisting of a binary target object modal mask and a two-channel depth image, is the only input used for training for seamless transfer from simulation to real images. The ground truth target object distribution is generated by summing all shifted amodal target object masks whose modal masks correspond with the target object modal mask.}
\label{fig:datagen}
\vspace{-8pt}
\end{figure*}

\subsection{Definitions} \label{subsec:defs}
We define the problem as a partially-observable Markov decision process (POMDP) with the 7-tuple $(S, A, T, R, \Omega, O, \gamma)$ and a maximum horizon $H$:
\begin{itemize}
    \item \textbf{States} $(S)$: A state $\mathbf{s}_k$ at timestep $k$ consists of the robot, a static overhead RGBD camera, and a static bin containing $N+1$ objects, target object $\mathcal{O}_t$ and distractor objects $\lbrace \mathcal{O}_{1,k}, \mathcal{O}_{2,k}, \ldots, \mathcal{O}_{N,k}\rbrace$. No prior information is known about the $N$ distractor objects. 
    \item \textbf{Actions} $(A)$: A grasp action $\mathbf{a}_k$ at timestep $k$ executed by the robot's gripper.
    \item \textbf{Transitions} $(T)$: In simulation, the transition model $T(\mathbf{s}_{k+1} \ | \ \mathbf{a}_k, \mathbf{s}_k)$ is equivalent to that used by Mahler et al.~\cite{mahler2017learning} and uses pybullet~\cite{coumans2017bullet} for dynamics. On the physical system, next states are determined by executing the action on a physical robot and waiting until objects come to rest.
    \item \textbf{Rewards} $(R)$: The reward $r_k = R(\mathbf{s}_k, \mathbf{a}_k, \mathbf{s}_{k+1}) \in \lbrace 0, 1 \rbrace$ is 1 if the target object is successfully grasped and lifted from the bin, otherwise the reward is 0.
    \item \textbf{Observations} $(\Omega)$: An observation $\mathbf{y}_k \in \mathbb{R}_+^{h \times w \times 4}$ at timestep $k$ consists of an RGBD image with width $w$ and height $h$ taken by the overhead camera.
    \item \textbf{Observation Model} $(O)$: A deterministic observation model $O(\mathbf{y}_k \ | \ \mathbf{s}_k)$ is defined by known camera intrinsics and extrinsics.
    \item \textbf{Discount Factor} $(\gamma)$: To encourage efficient extraction of the target object, $0 < \gamma < 1$.
\end{itemize}

We also define the following terms:
\begin{itemize}
    \item Modal Segmentation Mask $(\mathcal{M}_{m,i})$: the region(s) of pixels in an image corresponding to object $\mathcal{O}_i$ which are visible~\cite{kanizsa1979organization}.
    \item Amodal Segmentation Mask $(\mathcal{M}_{a,i})$: the region(s) of pixels in an image corresponding to object $\mathcal{O}_i$ which are visible or invisible (occluded by other objects in the image)~\cite{kanizsa1979organization}.
    \item The oriented minimum bounding box is the 3D box with the minimum volume that encloses the object, subject to no orientation constraints. We use this box to determine scale and aspect ratio for a target object.
    \item The \textit{occupancy distribution} $\rho \in \mathcal{P}$ is the unnormalized distribution describing the likelihood that a given pixel in the observation image contains some part of the target object's amodal segmentation mask.
\end{itemize}

\subsection{Objective} \label{subsec:objective}
Given this problem definition and assumptions, the objective is to find a policy $\pi_\theta^*$ with parameters $\theta$ that maximizes the expected discounted sum of rewards:
\begin{align*}
    \theta^* = \arg \max_\theta \ \mathbb{E}_{p(\tau | \theta)} \left[\sum_{k=0}^{H-1} \gamma^k R(\mathbf{s}_k, \pi_\theta(\mathbf{y}_k), \mathbf{s}_{k+1}) \right]
\end{align*}
where $p(\tau \ | \ \theta) = \mathbb{P}(s_0) \Pi_{k=0}^{H-1} T(\mathbf{s}_{k+1} \ | \ \pi_\theta(\mathbf{y}_k), \mathbf{s}_k) O(\mathbf{y}_k \ | \ \mathbf{s}_k)$ is the distribution of state trajectories $\tau$ induced by a policy $\pi_\theta$~\cite{mahler2017learning}. Maximizing this objective corresponds to removing the target object in the fewest number of actions.

\subsection{Surrogate Reward} \label{subsec:surrogate-reward}
Because the reward defined in Section~\ref{subsec:defs} is sparse and the transition function relies on complex inter-object and grasp contact dynamics, it is difficult to directly optimize for $\pi_\theta$. Thus, we instead introduce a dense surrogate reward $\Tilde{R}$ describing the reduction of the support of the target object's occupancy distribution:
\begin{align*}
    \Tilde{R}(\mathbf{y}_k, \mathbf{y}_{k+1}) = |\textrm{supp}(f_\rho(\mathbf{y}_{k}))| - |\textrm{supp}(f_\rho(\mathbf{y}_{k+1}))|,
\end{align*}
where $f_\rho : \Omega \longrightarrow \mathcal{P}$ is a function that takes an observation $\mathbf{y}_k$ and produces the corresponding occupancy distribution $\rho_k$ for a given bounding box and $\textrm{supp}(\rho) = \lbrace (i, j) \in \lbrace 0, \ldots, h-1 \rbrace \times \lbrace 0, \ldots, w-1 \rbrace \ | \ \rho(i,j) \neq 0$ is the \textit{support} of the occupancy distribution. Then, $|\textrm{supp}(\rho)|$ is the number of nonzero pixels in $\rho$. Section~\ref{sec:perception} discusses a data-driven approximation for the function $f_\rho$ while Section~\ref{sec:xray-policy} discusses a greedy policy using the learned $f_\rho$ and $\Tilde{R}$.
\section{Learning Occupancy Distributions} \label{sec:perception}
We describe a method for estimating the function $f_\rho$ via a deep neural network.  Each pixel in the occupancy distribution $\rho \in [0, 1]^{h \times w}$ has a value representing the likelihood of it containing part of the target object's amodal segmentation mask, or the likelihood that some part of the object, in some planar translation or rotation, would occupy that pixel without any occlusions from other objects. We train this pixelwise distribution network on a dataset of augmented depth images and ground-truth occupancy distributions.

\begin{table*}[th!]
	\centering
	\vspace{1.5mm}
	\begin{tabu} to \textwidth {X[2c]X[c]X[c]X[c]X[c]X[c]X[c]X[c]X[c]} \toprule
		  & \multicolumn{2}{c}{\textbf{Test}} & \multicolumn{2}{c}{\textbf{Lid}} & \multicolumn{2}{c}{\textbf{Domino}} & \multicolumn{2}{c}{\textbf{Flute}}  \\
		\textbf{Aspect Ratio} & Bal. Acc. & IoU & Bal. Acc. & IoU & Bal. Acc. & IoU & Bal. Acc. & IoU \\\midrule
		1:1 & $98\%$ & $0.91$ & $\bm{93\%}$ & $\bm{0.70}$ & $92\%$ & $0.74$ & $71\%$ & $0.30$\\
		2:1 & $97\%$ & $0.90$ & $79\%$ & $0.44$ & $\bm{96\%}$ & $0.81$ & $84\%$ & $0.44$\\
		5:1 & $97\%$ & $0.90$ & $66\%$ & $0.23$ & $96\%$ & $\bm{0.83}$ & $\bm{86\%}$ & $\bm{0.49}$\\
		10:1 & $97\%$ & $0.87$ & $84\%$ & $0.49$ & $82\%$ & $0.58$ & $82\%$ & $0.41$ \\\bottomrule
	\end{tabu}
	\caption{Balanced accuracy (Bal. Acc.) and Intersection over Union (IoU) metrics for networks trained on various aspect ratio target boxes. The first column is the respective set of 2,000 test images for the network's training dataset. The other columns show how the networks can generalize to unseen objects outside the training distribution. Each dataset contains 1,000 test images for the lid, domino, and flute objects, respectively. These objects are shown in Figure~\ref{fig:perceptionbenchmark} and have approximate aspect ratios of 1:1, 2:1, and 5:1, respectively. Each network performs very well when estimating distributions for its training target object and makes reasonable predictions for target objects with similar bounding box aspect ratios, even for novel target objects at different scales and in the presence of new occluding objects. However, a network trained on a small aspect ratio does not generalize well to higher aspect ratio objects, as it tends to overestimate the occupancy distribution.}
	\label{tab:perceptionbenchmark}
	\vspace{-6pt}
\end{table*}

\subsection{Dataset Generation} We generate a dataset of 10,000 synthetic augmented depth images labeled with target object occupancy distributions for a rectangular box target object. We choose 10 box targets of various dimensions ranging from $3 cm \times 3 cm \times 5 mm$ to $9.5 cm \times 0.95 cm \times 5 mm$ (aspect ratios varying from 1:1 to 10:1) with equal volume and generate a dataset for each, resulting in a total of 100,000 dataset images. We choose a relatively small thickness for the target so that it is more likely to be occluded in heaps of objects, as it tends to lie flat on the workspace. We sample a state $\mathbf{s}_0$ by uniformly sampling a set of $N$ 3D CAD models as well as a heap center and 2D offsets for each object from a 2D truncated gaussian. First, $\mathcal{O}_t$ is dropped from a fixed height above the workspace, then the other $N$ objects are dropped one by one from a fixed height and dynamic simulation is run until all objects come to rest (all velocities are zero). Any objects that fall outside of the workspace are removed. $N$ is drawn from a Poisson distribution ($\lambda = 12$) truncated such that $N \in [10, 15]$. The 3D CAD models are drawn from a dataset of 1296 models available on Thingiverse, including ``packaged" models, where the original model has been augmented with a rectangular backing, as in~\cite{mahler2019learning}. The camera position is drawn from a uniform distribution over a viewsphere and camera intrinsics are sampled uniformly from a range around their nominal values. We use the Photoneo Phoxi S datasheet intrinsics and a camera pose where the camera points straight down at the heap at a height of $0.8 m$ for the nominal values. An RGBD image is rendered and augmented depth images are created by concatenating a binary modal mask of the target object with the depth image. Note that if the target object is not visible, the image is equivalent to a two-channel depth image, as the first channel is all zeros. We find that training on these images, as opposed to training on RGBD images directly, allows for seamless transfer between simulated and real images.

To generate the ground-truth occupancy distribution, we find the set of translations and rotations in the image plane for the target object such that an image rendered from the same camera pose with all other objects in the scene in the same respective poses will yield the same target object modal segmentation mask. Thus, when the object is fully visible, the distribution's support collapses to the pixels of the target object modal segmentation mask. However, when the object is partially or fully occluded, then multiple target object translations or rotations may result in the same image and the distribution will spread to reflect where the target could hypothetically be hiding. In practice, we generate this distribution by discretizing the set of possible translations into a $64 \times 48$ grid (every 8 pixels in the image) and rotations into 16 bins, then shifting and rotating a target-only depth image to each point on the grid, offsetting by the depth of the bottom of the workspace at that point. By comparing the depths for the set of these shifted and rotated depth images to original depth image, we can determine the modal segmentation mask for the target object as if it were at each location. Any location for which there is intersection-over-union (IoU) greater than 0.9 (or, in cases where the target object has a blank modal mask due to full occlusion, any location for which the modal mask is also blank) is considered to result in the same image. Then, the amodal target object masks from all locations resulting in the same image are summed and the resulting normalized single-channel image is the ground truth occupancy distribution. A visualization of this process is shown in Figure~\ref{fig:datagen}. Dataset generation for 10,000 images took about 5 hours on an Ubuntu 16.04 machine with a 12-core 3.7 GHz i7-8700k processor.

\subsection{Occupancy Distribution Model} We split each dataset of 10,000 images image-wise and object-wise into training and test sets (8,000 training images and 2,000 test images, where objects are also split such that training objects only appear in training images and test objects only appear in test images). We train a fully-convolutional network with a ResNet-50 backbone~\cite{long2015fully} using a pixelwise mean-squared-error loss for 40 epochs with a learning rate of $10^{-5}$, momentum of 0.99, and weight decay of 0.0005. The input images were preprocessed by subtracting the mean pixel values calculated over the dataset and transposing to BGR. Training took approximately 2.5 hours on an NVIDIA V100 GPU and a single forward pass took 6 ms on average as compared to 1.5 s for generating the ground-truth distribution.

\subsection{Simulation Experiments for Occupancy Distributions}
\begin{figure}
    \centering
    \vspace{1.5mm}
    \includegraphics[width=\linewidth]{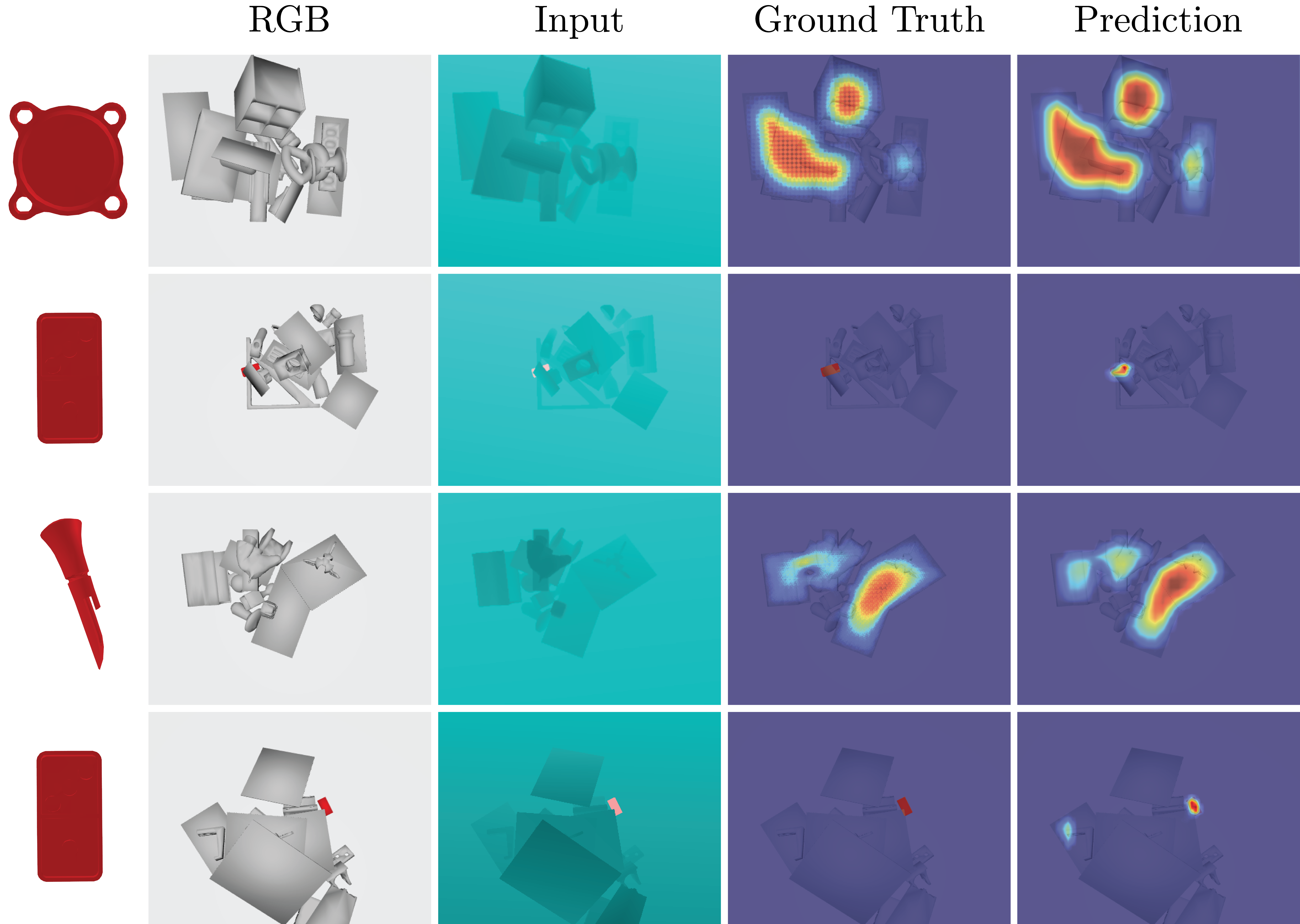}
    \caption{Example predicted target object occupancy distributions for three target objects, a lid, domino, and flute, unseen during training (far left). Warmer colors indicate a higher likelihood of that pixel containing part of the target object's amodal mask. The network is able to accurately predict a distribution across many objects, a collapsed distribution when the object is partially visible, and multimodal distributions when there are gaps between objects (top three rows). The final row shows a failure mode where the network spuriously predicts an extra mode for the distribution when the target object is partially occluded.}
    \label{fig:perceptionbenchmark}
    \vspace{-8pt}
\end{figure}

We benchmark the trained model on the full set of 2,000 test images as well as on 1,000 images with three other simulated target objects shown in Figure~\ref{fig:perceptionbenchmark} - a lid, a domino, and a flute - to test generalization to object shapes, aspect ratios and scales not seen during training. We chose these target objects due to their diversity in scale and object aspect ratio (e.g., the flute is longer, thinner, and deeper, while the lid is nearly square and flat). We report two metrics: balanced accuracy, the mean of pixelwise accuracies on positive and negative pixel labels, and intersection-over-union, the sum of positive pixels in both the ground truth and predicted distribution divided by the sum of total positive pixels in either distribution. We consider true positives as the ground truth pixel having normalized value greater than 0.1 and the predicted value being within 0.2 of the ground truth value. Similarly, we consider true negatives as the ground truth pixel having normalized value less than 0.1 and the predicted value being within 0.2 of the ground truth value. Results are shown in Table~\ref{tab:perceptionbenchmark}. 

\textbf{Target Object Scale.} For objects of different scale than the training target object, we scale the input image by a factor equal to the difference in scale between the box target object and the other target object, feed it through the network, and then rescale the output distribution. We find that this scaling dramatically improves performance with minimal preprocessing of the input image; for example, when testing on the lid object, which is about twice as large as the training box object, we increase balanced accuracy and IoU from $63.0\%$ and $0.186$ to $93.1\%$ and $0.697$, respectively.

\begin{figure}[t!]
    \centering
    \vspace{1.5mm}
    \includegraphics[width=0.75\linewidth]{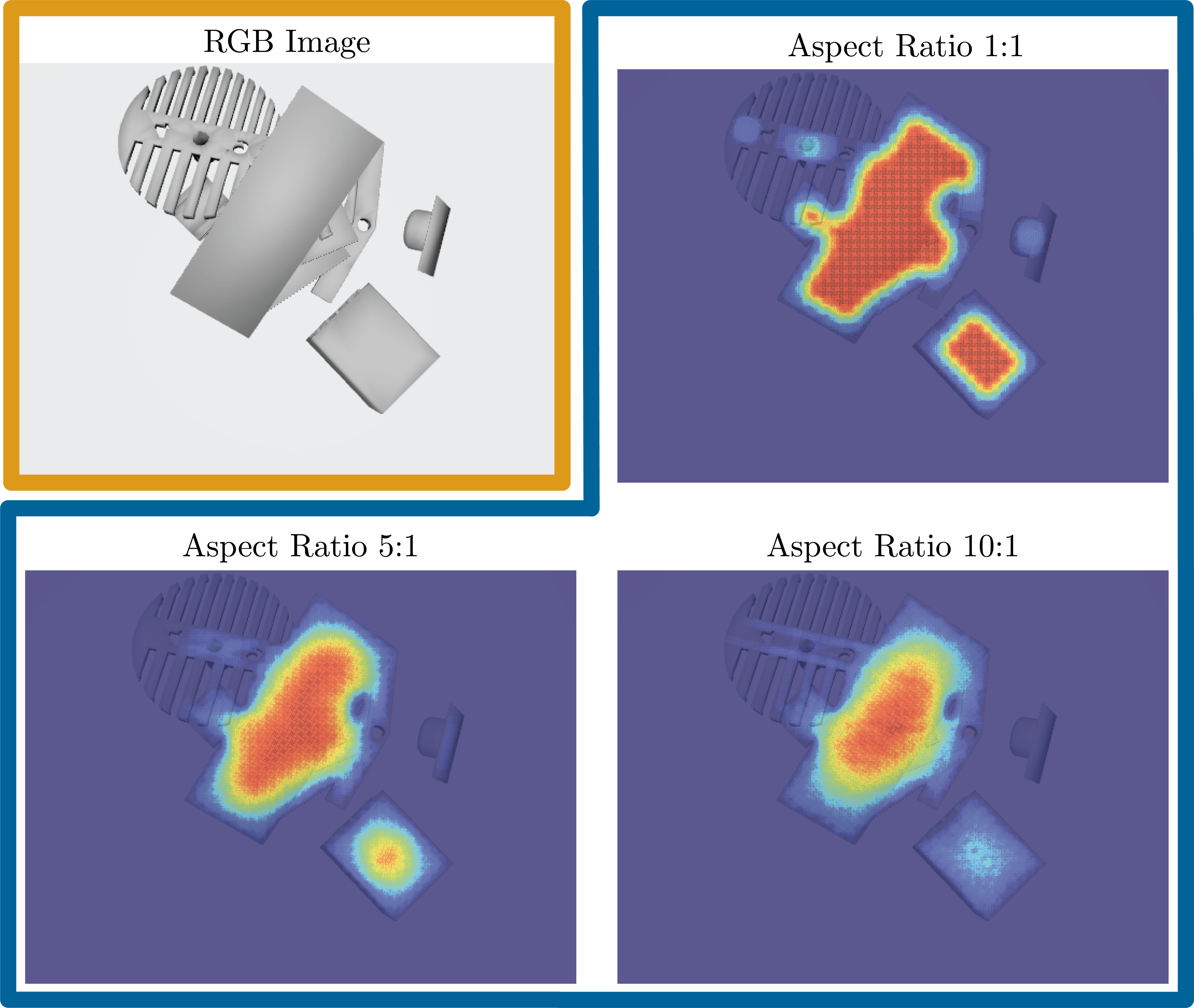}
    \caption{The ground truth occupancy distributions for a target object of various aspect ratios for the same heap image.}
    \label{fig:aspectratios}
    \vspace{-8pt}
\end{figure}

\textbf{Target Aspect Ratios.} We found that, while our network performed well on objects with similar aspect ratios, longer and thinner objects with higher aspect ratios resulted in the model overestimating the support of the distribution. This effect can be seen in Figure~\ref{fig:aspectratios}, which shows ground truth occupancy distributions for target objects of different aspect ratios in the same heap image. Table~\ref{tab:perceptionbenchmark} suggests that the trained networks can accurately predict occupancy distributions for target objects that have similar aspect ratios to the training boxes, but do not perform as well when tasked with predicting a distribution for objects with dramatically different aspect ratios. In particular, the network trained with a 1:1 box target object tends to overestimate the support for target objects with high aspect ratios, leading to a drop in metrics. This effect is especially visible along corners of occluding objects, where more rotations of a low aspect ratio object are possible, while only one or two rotations of a high aspect ratio object are possible.

\begin{figure*}[th!]
\centering
\vspace{1.5mm}
\includegraphics[width=\textwidth]{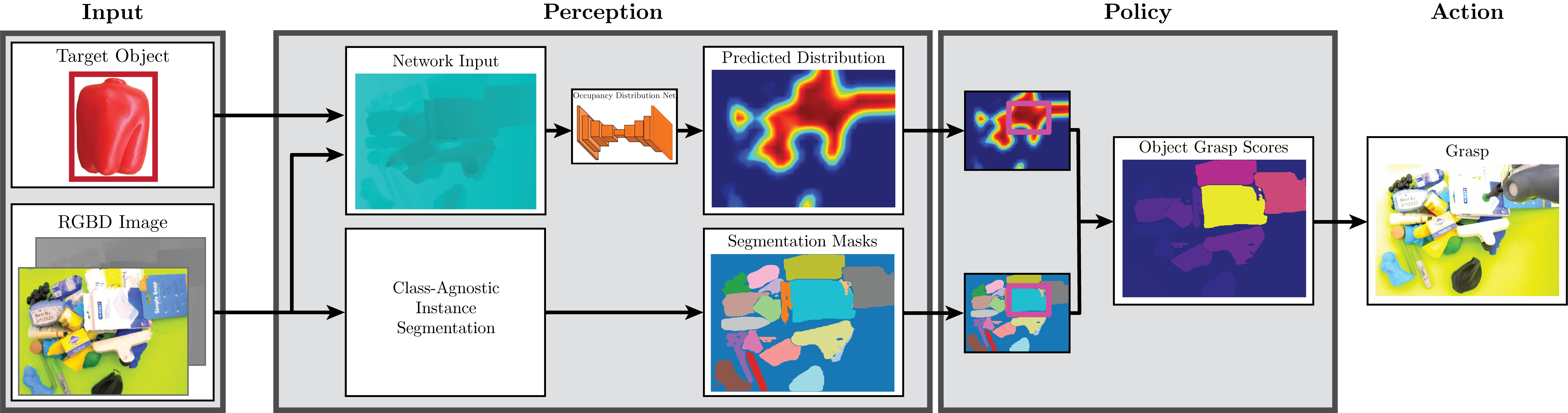}
\caption{The perception stage takes as input an RGBD image of the scene and outputs an occupancy distribution prediction using a network based on the target object bounding box dimensions and the created augmented depth image. The perception stage also produces a set of segmentation masks. The X-Ray mechanical search policy then finds the mask that has the most overlap with the occupancy distribution (colored yellow in the grasp scores image) and plans a grasp on that mask.}
\label{fig:policy}
\vspace{-8pt}
\end{figure*}

Figure~\ref{fig:perceptionbenchmark} shows occupancy distribution predictions with ground truth distributions for the three unseen objects using the network trained on the closest aspect ratio target object and scaled appropriately. Results suggest that the network is able to accurately predict diverse distributions when occluding objects not seen in training are present. Figure~\ref{fig:perceptionbenchmark} suggests not only that the network can predict the correct distribution spanning multiple occluding objects in unimodal and multimodal cases when the target object is fully occluded, but also that it can correctly collapse the distribution to a small area around the visible part of the target object when it is only partially occluded.

\section{X-Ray: Mechanical Search Policy} \label{sec:xray-policy}

Using the learned occupancy distribution function $f_\rho$, we propose X-Ray, a mechanical search policy that optimizes for the objective and surrogate reward $\Tilde{R}$ defined in Section~\ref{sec:problem}. We create both simulated and physical object heaps and generate overhead camera images using an observation model based on the Photoneo PhoXi S depth camera. The heap RGBD image and target object are inputs to the perception system, which uses the network trained on the most similar bounding box to the target object to predict an occupancy distribution for the target. The policy takes the predicted distribution and a set of modal segmentation masks for the scene and computes a grasping action that would maximally reduce the support of the subsequent distribution. Specifically, the policy takes an element-wise product of each segmentation mask with the predicted occupancy distribution and sums over all entries in the resulting image, leading to a score for each of the segmentation masks. The policy then plans a grasp on the object mask with the highest score and executes it, as shown in Figure~\ref{fig:policy}. 

\subsection{Simulation Experiments with X-Ray}
We first evaluate the mechanical search policy with simulated heaps of novel objects. To further test the ability of the learned network to generalize to unseen occluding objects, we use a set of objects unseen in training and validation: 46 YCB objects~\cite{calli2015benchmarking} and 13 ``packaged'' YCB objects (augmented in the same way as described in Section~\ref{sec:perception}). Initial states were generated as explained in Section~\ref{sec:perception}, first dropping the target object, followed by the other $N$ objects. We use $N=14$ so each heap initially contained 15 total objects, color{red}a similar or larger size to previous bin-picking work~\cite{mahler2017learning,morrison2018closing}. As the focus of this work was not instance segmentation or target detection, we use ground truth segmentation masks and target binary masks in simulation, although we note that any class-agnostic instance segmentation network~\cite{kuo2019shapemask,danielczuk2019segmenting} or object detection network~\cite{zhao2019object} can be substituted. For each grasp, either a parallel jaw or suction cup grasp, we use wrench space analysis to determine whether it would result in the object being lifted from the workspace under quasi-static conditions~\cite{prattichizzo2008grasping,mahler2016dex, mahler2017dex}. If the grasp is collision-free and the object can be lifted, the object is lifted until the remaining objects come to rest using dynamic simulation implemented in pybullet, resulting in the next state. Otherwise the state remains unchanged.

\begin{table}
	\centering
	\begin{tabu} to \linewidth {XX[2c]X[c]X[c]X[c]}
	\toprule
		\textbf{Policy} & \textbf{Success Rate} & \multicolumn{3}{c}{\textbf{Number of Actions Quartiles}} \\\midrule
		Random & $42\%$ & $4$ & $7$ & $9$ \\
		Largest & $67\%$ & $4$ & $\mathbf{5}$ & $7$ \\
		X-Ray & $\bm{82\%}$ & $\mathbf{3}$ & $\mathbf{5}$ & $\mathbf{6}$ \\\bottomrule
	\end{tabu}
	\caption{Evaluation metrics for each policy over 1,000 simulated rollouts. The lower quartiles, medians, and upper quartiles for number of actions are reported for successful rollouts. X-Ray extracts the target at a higher success rate with significantly fewer actions.}
	\label{tab:simresults}
	\vspace{-6pt}
\end{table}

In addition to the policy proposed here, we evaluate two previously proposed baseline policies, \textbf{Random} and \textbf{Largest}~\cite{danielczuk2019mechanical}. The \textbf{Random} policy that first attempts to grasp the target object, and, if no grasps are available on the target object, grasps an object chosen uniformly at random from the bin. The \textbf{Largest} policy that first attempts to grasp the target object, and, if no grasps are available on the target object, iteratively attempts to grasp the objects in the bin according to the size of their modal segmentation mask.

\begin{figure}
\centering
\includegraphics[width=0.95\linewidth]{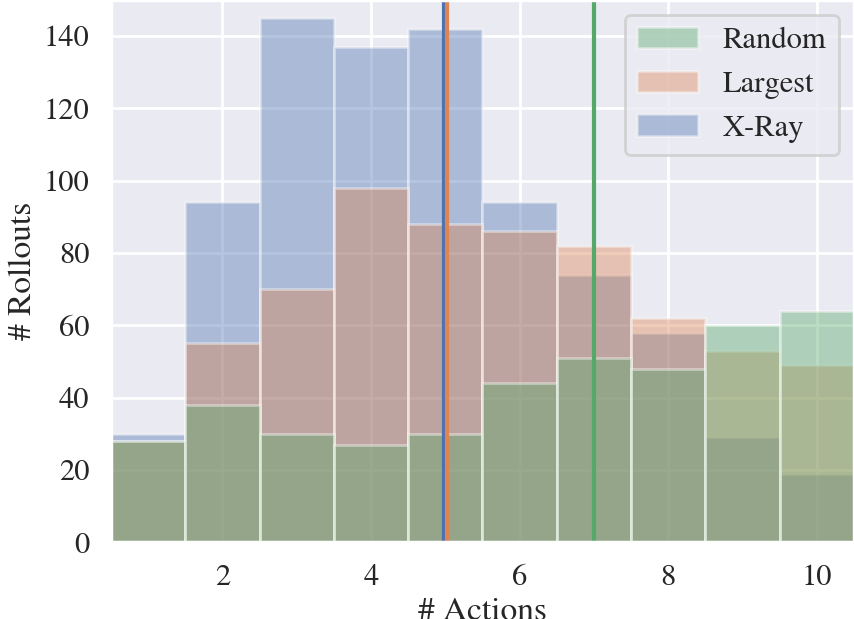}
\caption{Histogram of the number of actions taken to extract the target object over the 1,000 simulated rollouts for the three policies tested. The median number of actions for each policy is shown by the corresponding vertical line.}
\label{fig:simresults}
\vspace{-8pt}
\end{figure}

Each policy was rolled out on 1,000 total heaps until either the target object was grasped (successful rollout) or the horizon $H=10$  was reached (failed rollout). We benchmark each policy using two metrics: success rate of the policy and mean number of actions taken to extract the target object in successful rollouts. Table~\ref{tab:simresults} and Figure~\ref{fig:simresults} show these metrics and the distribution of successful rollouts over the number of actions taken to extract the target object, respectively.

While the Random and Largest policies occasionally are able to quickly extract the target object, X-Ray consistently extracts the target in fewer actions and succeeds in 15\% more heaps than the best-performing baseline. Largest is a reasonable heuristic for these heaps, as shown in~\cite{danielczuk2019mechanical}, as large objects typically have a greater chance of occluding the target, but X-Ray combines this intuition with superior performance when the object is partially occluded. X-Ray outperforms the Largest policy on heaps where the target object is partially occluded by a thin or small object (such as a fork or dice) at some point during the rollout. In these scenarios, a robust grasp is often not available on  the target object, and while X-Ray can correctly identify that the occluding object should be removed, the Largest policy will often grasp a larger object further from the target object. In scenarios where there are many large objects, but some are lying to the side, X-Ray will typically grasp objects that are in the more cluttered area of the bin, since they are more likely to reveal the target object. This behavior is a function of weighting the object area by the predicted distribution, which encourages the policy to ignore solitary objects.

\subsection{Physical Experiments with X-Ray}
We also evaluate X-Ray with heaps of novel household objects on a physical ABB YuMi robot with a suction cup and parallel jaw gripper, using two target objects. Some examples of the objects used can be seen in Figures~\ref{fig:splash} and~\ref{fig:policy}. Initial states were generated by placing the target object on the workspace, filling a bin with the $N$ other objects, and then dumping the bin on top of the target object. In these heaps, $N=24$ was used so that each heap initially contained 25 total objects. We chose 25 total objects because it has been commonly used in cluttered bin-picking environments~\cite{mahler2019learning} and objects tend to disperse further on the physical setup. For segmentation masks, we used the class-agnostic instance segmentation network from~\cite{danielczuk2019segmenting}, and for grasp quality analysis, we used FC-GQCNN~\cite{satish2019policy}. To generate binary target masks, we use HSV color segmentation from OpenCV and use red target objects. While we make this assumption for simplicity, we note that we could substitute this process with a target object segmentation method that uses visual features, semantics and shape, such as the one described in~\cite{danielczuk2019segmenting}.

We perform 20 rollouts for each of the three policies. Each policy was rolled out until either the target object was grasped (successful rollout) or the horizon $H=10$  was reached (failed rollout). We report the same metrics as in the simulated experiments in Table~\ref{tab:physicalresults}.

We find that X-Ray outperforms both baselines, extracting the target object in a median 5 actions over the 20 rollouts as compared to 6 actions for the Largest and Random policies while succeeding in extracting the target object within 10 actions in each case. These results suggest that X-Ray not only can extract the target more efficiently than the baseline policies, but also has lower variance. The Largest policy performed comparatively worse with more objects in the heap than in simulation, as it relies heavily on accurate segmentation masks. However, when objects are densely clustered together, segmentation masks are often merged, leading to grasps on smaller objects that do not uncover the target. In this case or in the case of spurious segmentation masks that do not cover objects, X-Ray reduces this reliance on accurate segmentation masks, as the occupancy distribution and segmentation are combined to create a score for the mask. This property of X-Ray causes it to compare favorably to a policy that directly scores segmentation masks based on their relationship to the target object geometry. X-Ray also reduces reliance on the target object binary mask being accurate; if the detector cannot see enough of the target object to generate a detection even when it is partially visible, X-Ray will continue to try and uncover it according to the fully occluded occupancy distribution until more of the target is revealed.

\begin{table}
\vspace{2mm}
	\centering
	\begin{tabu} to \linewidth {XX[2c]X[c]X[c]X[c]} \toprule
		\textbf{Policy} &\textbf{Success Rate} &  \multicolumn{3}{c}{\textbf{Number of Actions Quartiles}} \\\midrule
		Random & $85\%$ & $\mathbf{4}$ & $6$ & $7$  \\
		Largest & $85\%$ & $4$ & $6$ & $7$ \\
		X-Ray & $\bm{100\%}$ & $\mathbf{4}$ & $\mathbf{5}$ & $\mathbf{5.25}$ \\\bottomrule
	\end{tabu}
	\caption{Evaluation metrics for each policy over 20 physical rollouts. The lower quartiles, medians, and upper quartiles for the number of actions are reported across successful rollouts. X-Ray extracts the target with significantly fewer actions, always extracting it within 10 actions.}
	\label{tab:physicalresults}
	\vspace{-6pt}
\end{table}


\section{Discussion and Future Work} \label{sec:discussion}
We present X-Ray, a mechanical search algorithm that minimizes support of a learned occupancy distribution. We showed that a model trained only on a synthetic dataset of augmented depth images labeled with ground truth distributions learns to accurately predict occupancy distributions for target objects unseen in training. We benchmark X-Ray in both simulated and physical experiments, showing that it can efficiently extract the target object from challenging heaps containing 15-25 objects that fully occlude the target object in 82\% - 100\% of heaps using a median of just 5 actions.

In future work, we will address some of the failure modes of the system, especially for objects that are significantly non-planar. Currently, the assumption that the object is flat can result in incorrect occupancy distributions for taller objects. Additionally, we will look to add memory to the policy so that if objects shift into previously free space, the distribution will not cover that area, and explore reinforcement learning policies based on a reward of target object visibility.

\section*{Acknowledgments}
\footnotesize
This research was performed at the AUTOLAB at UC Berkeley in affiliation with the Berkeley AI Research (BAIR) Lab. The authors were supported in part by donations from Google. This material is based upon work supported by the National Science Foundation Graduate Research Fellowship Program under Grant No. 1752814. Any opinions, findings, and conclusions or recommendations expressed in this material are those of the author(s) and do not necessarily reflect the views of the Sponsors. We thank our colleagues and collaborators who provided helpful feedback, code, and suggestions, especially Julian Ibarz, Brijen Thananjeyan, Andrew Li, Andrew Lee, Andrey Kurenkov, Roberto Mart\'in Mart\'in, Animesh Garg, Matt Matl, and Ashwin Balakrishna.

\renewcommand*{\bibfont}{\footnotesize}
\clearpage
\printbibliography 

@STRING{icra = {{Proc. {IEEE} Int. Conf. Robotics and Automation (ICRA)}}}

@STRING{iros = {Proc. IEEE/RSJ Int. Conf. on Intelligent Robots and Systems (IROS)}}

@STRING{iccv = {Proc. {IEEE} Int. Conf. on Computer Vision (ICCV)}}

@STRING{eccv = {Proc. European Conf. on Computer Vision (ECCV)}}

@STRING{accv = {Proc. Asian Conf. on Computer Vision (ACCV)}}

@STRING{cvpr = {Proc. {IEEE} Conf. on Computer Vision  and Pattern Recognition (CVPR)}}

@STRING{humanoids = {Proc. {IEEE-RAS} Int. Conf. on Humanoid Robots}}

@STRING{neurips = {Proc. Advances in Neural Information Processing Systems}}

@STRING{isrr = {Int. S. Robotics Research (ISRR)}}

@STRING{ijrr = {Int. Journal of Robotics Research (IJRR)}}

@STRING{rss = {Proc. Robotics: Science and Systems (RSS)}}

@STRING{ral = {IEEE Robotics \& Automation Letters}}

@STRING{corl = {Conf. on Robot Learning (CoRL)}}

@inproceedings{manhardt2018explaining,
  title={Explaining the Ambiguity of Object Detection and 6D Pose from Visual Data},
  author={Manhardt, Fabian and Arroyo, Diego Martin and Rupprecht, Christian and Busam, Benjamin and Navab, Nassir and Tombari, Federico},
  booktitle=iccv,
  year={2019}
}

@article{mahler2017dex,
  title={Dex-Net 2.0: Deep Learning to Plan Robust Grasps with Synthetic Point Clouds and Analytic Grasp Metrics},
  author={Mahler, Jeffrey and Liang, Jacky and Niyaz, Sherdil and Laskey, Michael and Doan, Richard and Liu, Xinyu and Ojea, Juan Aparicio and Goldberg, Ken},
  journal=rss,
  year={2017},
}

@incollection{prattichizzo2008grasping,
  title={Grasping},
  author={Prattichizzo, Domenico and Trinkle, Jeffrey C},
  booktitle={Springer handbook of robotics},
  pages={671--700},
  year={2008},
  publisher={Springer}
}

@MISC{coumans2017bullet,
    author =   {Erwin Coumans and Yunfei Bai},
    title =    {pybullet, a Python module for physics simulation, games, robotics and machine learning},
    howpublished = {\url{http://pybullet.org/}},
    year = {2017}
}

@inproceedings{mahler2017learning,
  title={Learning deep policies for robot bin picking by simulating robust grasping sequences},
  author={Mahler, Jeffrey and Goldberg, Ken},
  booktitle=corl,
  pages={515--524},
  year={2017}
}

@book{kanizsa1979organization,
  title={Organization in vision: Essays on Gestalt perception},
  author={Kanizsa, Gaetano},
  year={1979},
  publisher={Praeger Publishers}
}

@inproceedings{long2015fully,
  title={Fully convolutional networks for semantic segmentation},
  author={Long, Jonathan and Shelhamer, Evan and Darrell, Trevor},
  booktitle=cvpr,
  pages={3431--3440},
  year={2015}
}

@inproceedings{danielczuk2019segmenting,
  title={Segmenting unknown 3d objects from real depth images using mask r-cnn trained on synthetic data},
  author={Danielczuk, Michael and Matl, Matthew and Gupta, Saurabh and Li, Andrew and Lee, Andrew and Mahler, Jeffrey and Goldberg, Ken},
  booktitle=icra,
  pages={7283--7290},
  year={2019},
  organization={IEEE}
}

@article{kuo2019shapemask,
  title={ShapeMask: Learning to Segment Novel Objects by Refining Shape Priors},
  author={Kuo, Weicheng and Angelova, Anelia and Malik, Jitendra and Lin, Tsung-Yi},
  journal={arXiv preprint arXiv:1904.03239},
  year={2019}
}

@inproceedings{price2019inferring,
  title={Inferring occluded geometry improves performance when retrieving an object from dense clutter},
  author={Price, Andrew and Jin, Linyi and Berenson, Dmitry},
  booktitle=isrr,
  year={2019}
}

@inproceedings{prokudin2018deep,
  title={Deep directional statistics: Pose estimation with uncertainty quantification},
  author={Prokudin, Sergey and Gehler, Peter and Nowozin, Sebastian},
  booktitle=eccv,
  pages={534--551},
  year={2018}
}

@inproceedings{rupprecht2017learning,
  title={Learning in an uncertain world: Representing ambiguity through multiple hypotheses},
  author={Rupprecht, Christian and Laina, Iro and DiPietro, Robert and Baust, Maximilian and Tombari, Federico and Navab, Nassir and Hager, Gregory D},
  booktitle=iccv,
  pages={3591--3600},
  year={2017}
}

@inproceedings{kehl2017ssd,
  title={SSD-6D: Making RGB-based 3D detection and 6D pose estimation great again},
  author={Kehl, Wadim and Manhardt, Fabian and Tombari, Federico and Ilic, Slobodan and Navab, Nassir},
  booktitle=iccv,
  pages={1521--1529},
  year={2017}
}

@article{xiang2017posecnn,
  title={Posecnn: A convolutional neural network for 6d object pose estimation in cluttered scenes},
  author={Xiang, Yu and Schmidt, Tanner and Narayanan, Venkatraman and Fox, Dieter},
  journal={arXiv preprint arXiv:1711.00199},
  year={2017}
}

@inproceedings{li2018deepim,
  title={Deepim: Deep iterative matching for 6d pose estimation},
  author={Li, Yi and Wang, Gu and Ji, Xiangyang and Xiang, Yu and Fox, Dieter},
  booktitle=eccv,
  pages={683--698},
  year={2018}
}

@inproceedings{hinterstoisser2012model,
  title={Model based training, detection and pose estimation of texture-less 3d objects in heavily cluttered scenes},
  author={Hinterstoisser, Stefan and Lepetit, Vincent and Ilic, Slobodan and Holzer, Stefan and Bradski, Gary and Konolige, Kurt and Navab, Nassir},
  booktitle=accv,
  pages={548--562},
  year={2012},
  organization={Springer}
}

@inproceedings{pinto2016supersizing,
  title={Supersizing self-supervision: Learning to grasp from 50k tries and 700 robot hours},
  author={Pinto, Lerrel and Gupta, Abhinav},
  booktitle=icra,
  pages={3406--3413},
  year={2016},
  organization={IEEE}
}

@article{kalashnikov2018qt,
  title={Qt-opt: Scalable deep reinforcement learning for vision-based robotic manipulation},
  author={Kalashnikov, Dmitry and Irpan, Alex and Pastor, Peter and Ibarz, Julian and Herzog, Alexander and Jang, Eric and Quillen, Deirdre and Holly, Ethan and Kalakrishnan, Mrinal and Vanhoucke, Vincent and others},
  journal={arXiv preprint arXiv:1806.10293},
  year={2018}
}

@article{morrison2018closing,
  title={Closing the loop for robotic grasping: A real-time, generative grasp synthesis approach},
  author={Morrison, Douglas and Corke, Peter and Leitner, J{\"\u}rgen},
  journal={arXiv preprint arXiv:1804.05172},
  year={2018}
}

@article{yang2019deep,
  title={A Deep Learning Approach to Grasping the Invisible},
  author={Yang, Yang and Liang, Hengyue and Choi, Changhyun},
  journal={arXiv preprint arXiv:1909.04840},
  year={2019}
}

@inproceedings{zeng2018learning,
  title={Learning synergies between pushing and grasping with self-supervised deep reinforcement learning},
  author={Zeng, Andy and Song, Shuran and Welker, Stefan and Lee, Johnny and Rodriguez, Alberto and Funkhouser, Thomas},
  booktitle=iros,
  pages={4238--4245},
  year={2018},
  organization={IEEE}
}

@inproceedings{danielczuk2019mechanical,
  title={Mechanical Search: Multi-Step Retrieval of a Target Object Occluded by Clutter},
  author={Danielczuk, Michael and Kurenkov, Andrey and Balakrishna, Ashwin and Matl, Matthew and Wang, David and Mart{\'i}n-Mart{\'i}n, Roberto and Garg, Animesh and Savarese, Silvio and Goldberg, Ken},
  booktitle=icra,
  year={2019},
}

@inproceedings{varley2017shape,
  title={Shape completion enabled robotic grasping},
  author={Varley, Jacob and DeChant, Chad and Richardson, Adam and Ruales, Joaqu{\'\i}n and Allen, Peter},
  booktitle=iros,
  pages={2442--2447},
  year={2017},
  organization={IEEE}
}

@inproceedings{berenson2008grasp,
  title={Grasp synthesis in cluttered environments for dexterous hands},
  author={Berenson, Dmitry and Srinivasa, Siddhartha S},
  booktitle=humanoids,
  pages={189--196},
  year={2008},
  organization={IEEE}
}

@inproceedings{saxena2008learning,
  title={Learning grasp strategies with partial shape information.},
  author={Saxena, Ashutosh and Wong, Lawson LS and Ng, Andrew Y},
  booktitle={AAAI},
  volume={3},
  number={2},
  pages={1491--1494},
  year={2008}
}

@article{moll2017randomized,
  title={Randomized physics-based motion planning for grasping in cluttered and uncertain environments},
  author={Moll, Mark and Kavraki, Lydia and Rosell, Jan and others},
  journal=ral,
  volume={3},
  number={2},
  pages={712--719},
  year={2017},
  publisher={IEEE}
}

@inproceedings{mahler2016dex,
  title={Dex-net 1.0: A cloud-based network of 3d objects for robust grasp planning using a multi-armed bandit model with correlated rewards},
  author={Mahler, Jeffrey and Pokorny, Florian T and Hou, Brian and Roderick, Melrose and Laskey, Michael and Aubry, Mathieu and Kohlhoff, Kai and Kr{\"o}ger, Torsten and Kuffner, James and Goldberg, Ken},
  booktitle=icra,
  pages={1957--1964},
  year={2016},
  organization={IEEE}
}

@article{katz2014perceiving,
  title={Perceiving, learning, and exploiting object affordances for autonomous pile manipulation},
  author={Katz, Dov and Venkatraman, Arun and Kazemi, Moslem and Bagnell, J Andrew and Stentz, Anthony},
  journal={Autonomous Robots},
  volume={37},
  number={4},
  pages={369--382},
  year={2014},
  publisher={Springer}
}

@article{jang2017end,
  title={End-to-end learning of semantic grasping},
  author={Jang, Eric and Vijayanarasimhan, Sudheendra and Pastor, Peter and Ibarz, Julian and Levine, Sergey},
  journal={arXiv preprint arXiv:1707.01932},
  year={2017}
}

@article{lenz2015deep,
  title={Deep learning for detecting robotic grasps},
  author={Lenz, Ian and Lee, Honglak and Saxena, Ashutosh},
  journal=ijrr,
  volume={34},
  number={4-5},
  pages={705--724},
  year={2015},
  publisher={SAGE Publications Sage UK: London, England}
}

@inproceedings{xiao2019online,
  title={Online Planning for Target Object Search in Clutter under Partial Observability},
  author={Xiao, Yuchen and Katt, Sammie and ten Pas, Andreas and Chen, Shengjian and Amato, Christopher},
  booktitle=icra,
  pages={8241--8247},
  year={2019},
  organization={IEEE}
}

@article{calli2015benchmarking,
  title={Benchmarking in manipulation research: The YCB object and model set and benchmarking protocols},
  author={Calli, Berk and Walsman, Aaron and Singh, Arjun and Srinivasa, Siddhartha and Abbeel, Pieter and Dollar, Aaron M},
  journal={arXiv preprint arXiv:1502.03143},
  year={2015}
}

@article{mahler2019learning,
  title={Learning ambidextrous robot grasping policies},
  author={Mahler, Jeffrey and Matl, Matthew and Satish, Vishal and Danielczuk, Michael and DeRose, Bill and McKinley, Stephen and Goldberg, Ken},
  journal={Science Robotics},
  volume={4},
  number={26},
  pages={eaau4984},
  year={2019},
  publisher={Science Robotics}
}

@inproceedings{gualtieri2016high,
  title={High precision grasp pose detection in dense clutter},
  author={Gualtieri, Marcus and Ten Pas, Andreas and Saenko, Kate and Platt, Robert},
  booktitle=iros,
  pages={598--605},
  year={2016},
  organization={IEEE}
}

@inproceedings{rad2017bb8,
  title={Bb8: A scalable, accurate, robust to partial occlusion method for predicting the 3d poses of challenging objects without using depth},
  author={Rad, Mahdi and Lepetit, Vincent},
  booktitle=iccv,
  pages={3828--3836},
  year={2017}
}

@inproceedings{yang2019inferring,
  title={Inferring Distributions Over Depth from a Single Image},
  author={Yang, Gengshan and Hu, Peiyun and Ramanan, Deva},
  booktitle=iros,
  year={2019}
}

@inproceedings{kohl2018probabilistic,
  title={A probabilistic u-net for segmentation of ambiguous images},
  author={Kohl, Simon and Romera-Paredes, Bernardino and Meyer, Clemens and De Fauw, Jeffrey and Ledsam, Joseph R and Maier-Hein, Klaus and Eslami, SM Ali and Rezende, Danilo Jimenez and Ronneberger, Olaf},
  booktitle=neurips,
  pages={6965--6975},
  year={2018}
}

@inproceedings{corona2018pose,
  title={Pose estimation for objects with rotational symmetry},
  author={Corona, Enric and Kundu, Kaustav and Fidler, Sanja},
  booktitle=iros,
  pages={7215--7222},
  year={2018},
  organization={IEEE}
}

@article{satish2019policy,
  title={On-policy dataset synthesis for learning robot grasping policies using fully convolutional deep networks},
  author={Satish, Vishal and Mahler, Jeffrey and Goldberg, Ken},
  journal=ral,
  volume={4},
  number={2},
  pages={1357--1364},
  year={2019},
  publisher={IEEE}
}

@article{kostrikov2016end,
  title={End to end active perception},
  author={Kostrikov, Ilya and Erhan, Dumitru and Levine, Sergey},
  year={2016}
}

@article{zhao2019object,
  title={Object detection with deep learning: A review},
  author={Zhao, Zhong-Qiu and Zheng, Peng and Xu, Shou-tao and Wu, Xindong},
  journal={IEEE Trans. Neural Networks and Learning Systems},
  volume={30},
  number={11},
  pages={3212--3232},
  year={2019}
}



\end{document}